\documentclass[preprint,12pt]{elsarticle}

\usepackage{amssymb}
\usepackage{amsmath}
\usepackage{amsmath}
\usepackage[ruled,vlined]{algorithm2e}

\usepackage{graphicx,url}
\usepackage{booktabs}
\usepackage[utf8]{inputenc}
\usepackage{booktabs}  
\usepackage{tabularx} 
\usepackage{xcolor}
\usepackage{ulem}
\usepackage{hyperref}

\journal{Knowledge-Based Systems}

\begin{document}

\begin{frontmatter}



\title{Assessing the Impact of Retrieval-Augmented Generation on Algorithmic Problem-Solving}

\author[label1]{Gabriel Souza Baggio, Gabriel Machado Lunardi}
\author[label2]{Guilherme Medeiros Machado, Subhankar Maity, Faiza Belbachir}
\author[label3]{José Palazzo Moreira de Oliveira}

\affiliation[label1]{organization={Federal University of Santa Maria},
             addressline={Roraima Avenue, 1000},
             city={Santa Maria},
             postcode={97105-800},
             state={Rio Grande do Sul},
             country={Brazil}}

\affiliation[label2]{organization={LyRIDS lab, ECE Engineering School},
             addressline={10 Rue Sextius Michel},
             city={Paris},
             postcode={75015},
             state={Île-de-France},
             country={France}}

\affiliation[label3]{organization={Federal University of Rio Grande do Sul},
             addressline={Informatics Institute, Bento Gonçalves Avenue, 9500},
             city={Porto Alegre},
             postcode={91501-970},
             state={Rio Grande do Sul},
             country={Brazil}}

\begin{abstract}


Competitive programming problems pose distinctive challenges for Large Language Models, as they demand not only proficient code generation but also advanced algorithmic reasoning. This study presents a Retrieval-Augmented Generation approach specifically designed to enhance the capacity of Large Language Models to address such problems. Our approach integrates an embedding-based retrieval mechanism and a similarity search algorithm to identify analogous problems within a knowledge database, employing a dynamically adjusted threshold to mitigate retrieval noise. We conducted an evaluation  using the CodeContests dataset, focusing on the test split problems that exhibit similarity to instances from the training split. The experimental framework considered three conditions: (i) problems provided solely with their original descriptions, (ii) problems supplemented with randomly selected examples from the training split, and (iii) problems augmented with similar examples retrieved through the proposed method.
In addition, we incorporated a Min‑K membership inference analysis to examine whether the generated code solutions were influenced by potential memorization of training data within the evaluated Large Language Models. This complementary perspective enables a deeper understanding of model behavior by distinguishing genuine generalization through retrieval from implicit recall of pretraining samples.
Performance was assessed using both pass rates and the CodeForces Elo rating of the problems. The results provide empirical evidence that similarity-based retrieval substantially enhances Large Language Models performance in competitive programming scenarios, offering an scalable framework for improving code generation in algorithmic problem-solving contexts.

\end{abstract}





\end{frontmatter}



\section{Introduction}
\label{sec:introduction}

Large Language Models (LLMs) have demonstrated proficiency in solving generic problems across various professional domains \cite{LLMCodeGeneration, jimenez2024swebenchlanguagemodelsresolve, PerformanceLLMLeetCode}. Although some evaluation benchmarks indicate high performance --- such as on HumanEval, where GPT-4o \cite{openai2024gpt4ocard} and Claude-3.5-Sonnet \cite{anthropic2024claude35sonnet} achieved precision scores of 90\% and 92\%, respectively \cite{yu2024humanevalprombpppro} ---, other benchmarks show that LLMs exhibit limitations on competitive programming tasks \cite{chen2021evaluatinglargelanguagemodels, zheng2025livecodebenchproolympiadmedalists, hossain2025llmprosanalyzinglargelanguage, BenchmarkLeetcode}. For instance, in the analysis by Zheng et al. \cite{zheng2025livecodebenchproolympiadmedalists}, Claude-3.5-Sonnet achieved a 0\% precision rate on medium and hard problems and only 14.1\% on easy ones. In the same benchmark, even the best-performing LLMs, such as Gemini-2.5-Pro \cite{deepmind2025gemini25pro} and DeepSeek-R1 \cite{deepseekai2025deepseekr1incentivizingreasoningcapability}, could not solve a single hard problem.

These errors typically arise due to a lack of specific contextual knowledge and a limited understanding of underlying data structures, leading to hallucinations --- outputs that are syntactically invalid or logically inconsistent \cite{hallucination}. 
Under a deeper epistemological analysis, it becomes evident that Large Language Models (LLMs) operate on opinion-based trajectories—that is, on the formulation of statements not necessarily grounded in verifiable evidence, but rather derived from inferences or interpretations constructed within a limited context of knowledge. Hallucination, in this sense, may be understood as the generation of a perception or belief that, while internally coherent, does not correspond to empirical reality. In LLMs, this phenomenon arises because the model ``believes” it is producing a consistent and contextually appropriate response, even though such a response lacks factual support.
To mitigate this, Retrieval-Augmented Generation (RAG) \cite{RAGIntroduction, gao2023retrieval} has been proposed as a mechanism to ground the model’s outputs in relevant external knowledge by, with an Information Retrieval (IR) algorithm, retrieving contextually similar information from a structured corpus \cite{RAGIntroduction, gao2023retrieval}. This work investigates how RAG can be used to enhance the problem-solving capabilities of LLMs.


The 
proposition
~for applying RAG in this context is based on two key observations. First, studies indicate that problems within competitive programming repositories often share underlying algorithmic patterns and solution structures \cite{deng2025cpretdatasetbenchmarkmodel, dickey2025failureplagiarismdetectioncompetitive}. Second, the practice of providing relevant examples in the prompt (few-shot prompting) is an established technique for improving LLM performance on complex tasks \cite{openai_prompt_engineering_2025, ma2023fairnessguidedfewshotpromptinglarge}. Based on this,
the hypothesis is
~that a RAG system that retrieves previously solved problems with analogous characteristics to a target problem can provide the model with relevant context and reduce hallucination. 

To achieve this, we developed an Information Retrieval (IR) pipeline that employs embedding similarity with a dynamic threshold \cite{ReEffIRDinamicoIntroduzido}. We evaluated this system on the CodeContests dataset \cite{li2022competition}, which is divided into a training set of 13,328 problems and a test set of 165 problems. In addition to the problem title and statement, each sample includes metadatdynamig programming, bitwise operations). The training split served as the knowledge base from which we retrieved similar examples to augment the LLM prompts for solving the problems in the test set. The correctness of the generated solutions was then evaluated using the CodeForces \cite{codeforces} judge system, allowing us to measure the impact of the retrieved context on the model's performance.

\begin{figure}[!h]
\centering
\includegraphics[width=1\textwidth]{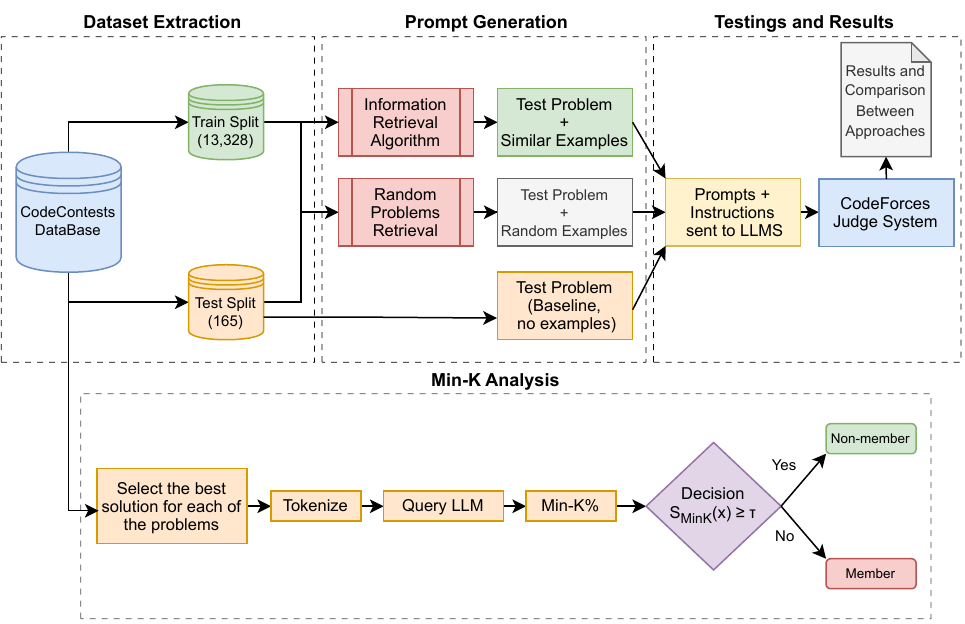}
\caption{System Architecture and Execution Pipeline}
\label{fig:flowchart}
\end{figure}

The evaluation involves three distinct prompting strategies: (i) a zero‑shot baseline using only the original description, (ii) a random retrieval baseline retrieving unrelated problems from the knowledge base, and (iii) the proposed similarity‑based RAG approach retrieving contextually aligned examples. Our analyses demonstrate that the similarity‑based retrieval consistently outperforms both baselines, validating the hypothesis that retrieving analogous problems improves LLM reasoning on complex programming tasks. Furthermore, the Min‑K membership inference results reveal that many generated code lines correspond to known training patterns, underscoring the importance of disentangling memorization effects from genuine reasoning improvements. Together, these findings provide a dual epistemic safeguard: RAG anchors outputs in external knowledge, while membership inference verifies their originality and data provenance. See Figure~\ref{fig:flowchart} for a graphical overview of the system architecture and execution pipeline.


The main contributions of this paper are summarized as follows:

\begin{enumerate}
    \item We design a RAG framework tailored to competitive programming tasks, integrating an embedding‑based similarity search with an adaptive thresholding mechanism to retrieve contextually relevant problems while minimizing retrieval noise.

    \item We construct a comprehensive experimental setup using the CodeContests dataset, evaluating six prominent LLMs across three prompting strategies, zero‑shot, random retrieval, and RAG‑based enhancement, to systematically quantify the effect of similarity‑driven retrieval on code generation performance.

    \item We introduce a complementary Min‑K Membership Inference analysis to assess model exposure and memorization biases, providing a novel perspective on the interplay between retrieval‑based augmentation and pretraining data recall.

    \item We present an integrated evaluation pipeline combining RAG effectiveness and membership inference insights, offering a dual perspective on both performance improvement and epistemic grounding in competitive programming contexts.

    \item Finally, we discuss the implications of these findings for the design of future reasoning‑oriented LLMs and outline directions for extending this work to upcoming state‑of‑the‑art models such as Gemini 3 and GPT‑5.
\end{enumerate}

\section{Related Works} 
\label{sec:related}

Several works have investigated the ability of LLMs to solve competitive programming problems, experimenting different evaluation metrics and approaches to maximize the correctness of generated solutions \cite{shi2024languagemodelssolveolympiad, quan2025codeelobenchmarkingcompetitionlevelcode, jain2024livecodebenchholisticcontaminationfree, zheng2025livecodebenchproolympiadmedalists, deng2025cpretdatasetbenchmarkmodel, hossain2025llmprosanalyzinglargelanguage, chen2021evaluatinglargelanguagemodels}. 

\subsection{Evaluation Strategies}

A main challenge in this domain is developing evaluation metrics that go beyond simple pass/fail rates and accurately reflect a model's problem-solving skill \cite{quan2025codeelobenchmarkingcompetitionlevelcode}. While simply annotating the number of correctly solved problems is informative, it fails to distinguish between solving a trivial problem and solving one that challenges even the best human competitors \cite{zheng2025livecodebenchproolympiadmedalists}.

To address this gap, recent benchmarks \cite{quan2025codeelobenchmarkingcompetitionlevelcode, openai2025introducing, boubdir2023elouncoveredrobustnessbest} have adopted the Elo rating system \cite{elo1978rating}, a method widely recognized for assessing the relative skill levels of players in competitive games --- and is also used in CodeForces platform. In the context of competitive programming platforms, the Elo system provides a unified scale to quantify both problem difficulty and competitor skill. The difficulty rating $d$ assigned to a problem is calibrated against the entire community of human participants; a contestant with an Elo rating of $d$ has a 50\% probability of solving this given problem \cite{boubdir2023elouncoveredrobustnessbest}.

The Elo rating of a contestant, in turn, provides a clear measure of their standing within the competitive community \cite{zheng2025livecodebenchproolympiadmedalists}. For example, a rating of 1600 (an ``Expert'' on CodeForces) places a user in the top 11\% of active participants, while a rating of 2400 (``Grandmaster'') signifies performance in the top 1\%. An elite rating of 2700 indicates a skill level superior to approximately 99.9\% of competitors. By framing LLM performance in this way, we can move from asking ``how many'' problems a model solved to ``how skilled'' the model is.

Following this evaluation method, our study treats each LLM as a virtual contestant and calculates its equivalent Elo rating. We specifically employ Bayesian Maximum-A-Posteriori (MAP) Elo rating \cite{zhou2024evaluatingsmoothcontrolattribute}, a method also utilized in \cite{zheng2025livecodebenchproolympiadmedalists}, to estimate the most probable skill level of a model given its performance.

The estimation process begins with the standard Elo probability function, $\pi_i(R)$, which calculates the likelihood of a model with rating $R$ correctly solving a problem with difficulty $d_i$:

\begin{equation*}
\pi_i(R) = \frac{1}{1 + 10^{(d_i - R)/400}}
\end{equation*}

These individual probabilities are used to construct the unnormalized log-posterior function, $\mathcal{L}(R)$. This function combines the log-likelihood of the observed outcomes ($y_i \in \{1 \text{ for accepted}, 0 \text{ for rejected}\}$) across all $N$ problems with a regularizing Gaussian prior on the model's rating (with mean $\mu_0$ and variance $\sigma_0^2$):

\begin{equation*}
\mathcal{L}(R) = \sum_{i=1}^{N} \left[ y_i \ln \pi_i(R) + (1-y_i) \ln(1 - \pi_i(R)) \right] - \frac{(R-\mu_0)^2}{2\sigma_0^2}
\end{equation*}

Finally, the MAP estimate, $\hat{R}_{\text{MAP}}$, is the rating $R$ that maximizes this log-posterior function:

\begin{equation*}
\hat{R}_{\text{MAP}} = \underset{R}{\arg\max} \ \mathcal{L}(R)
\end{equation*}

Since our benchmark utilizes the CodeContests dataset \cite{li2022competition}, which contains most of its problems from CodeForces, we annotated each problem's official Elo difficulty rating. This provides a standardized metric that allows for a comparison of our results against other benchmarks (assuming they also use Elo), even when the test sets are composed of different problems.

\subsection{Performance Enhancement Approaches}

Prior research has explored enhancing LLM performance through two main strategies: granting models access to external knowledge sources or fine-tuning them on specific datasets. For instance, augmenting models with tools like terminal and internet access has enabled systems such as o3 and o4-mini-high to achieve Elo ratings of 2706 and 2719 respectively \cite{openai2025introducing}. Another strategy involves fine-tuning LLMs on competitive programming datasets to train them on problem-solving patterns, a notable example being Google’s AlphaCode \cite{li2022competition}. Since these benchmarks also use Elo for evaluation, we used it as a common metric to assess the effectiveness of our technique in relation to these prior work.

\subsection{Problem Datasets}

Several datasets have been proposed to benchmark the ability of LLMs to solve competitive programming problems \cite{shi2024languagemodelssolveolympiad, quan2025codeelobenchmarkingcompetitionlevelcode, jain2024livecodebenchholisticcontaminationfree, zheng2025livecodebenchproolympiadmedalists, li2022competition}. Table~\ref{tab:datasets} summarizes these datasets, detailing how many problems each contains and the time period from which they were sourced.

\begin{table}[h]
\centering
\caption{Summary of competitive programming datasets used for LLM evaluation}
\label{tab:datasets}
\begin{tabular}{p{5cm} p{3cm} p{4cm}}
\toprule
\textbf{Dataset} & \textbf{\# Problems} & \textbf{Problems' Year(s)} \\
\midrule
USACO \cite{shi2024languagemodelssolveolympiad} & 307 & 2011-2024 \\
CodeElo \cite{quan2025codeelobenchmarkingcompetitionlevelcode} & 408 & 2024 \\
LiveCodeBench \cite{jain2024livecodebenchholisticcontaminationfree} & 442 & 2023-2024 \\
LiveCodeBench Pro \cite{zheng2025livecodebenchproolympiadmedalists} & $\sim$584 & Updated Lively \\
CodeContests \cite{li2022competition} & 13,493 & 2021 \\
\bottomrule
\end{tabular}
\end{table}

For this study, we selected the CodeContests dataset due to its extensive scale. Our central hypothesis is that retrieving similar, previously solved problems can enhance an LLM's performance. The large repository of over 13,000 problems in CodeContests provides a sufficiently dense knowledge base, which is essential for a retrieval-based approach to effectively find analogous examples.

Other benchmarks, while valuable, were less suited for our research objectives. The USACO dataset, for instance, sources its problems from a platform other than CodeForces, making it incompatible with the Elo rating system used for our evaluation. Datasets like CodeElo, LiveCodeBench, and LiveCodeBench Pro, despite their advantages of containing more recent problems and potentially having less data contamination, offer a very limited number of problems.

\section{Methodology}
\label{sec:methodology}

To evaluate RAG's impact on our work, we used six well-known LLMs and a set of programming questions, each paired with a previously tested and accepted solution. These entries serve as the knowledge base for the RAG. All datasets and code used in the methodology are included in our open-source repository\footnote{\label{fn:repo}\url{https://anonymous.4open.science/r/ArtigoSBBD-C51C}}.

\subsection{Model Selection}
\label{subsec:models}

The models utilized were chosen based on their use in prior research: DeepSeek-R1 \cite{zheng2025livecodebenchproolympiadmedalists, yu2024humanevalprombpppro, deng2025cpretdatasetbenchmarkmodel}, Gemini 2.0 Flash \cite{zheng2025livecodebenchproolympiadmedalists, jain2024livecodebenchholisticcontaminationfree, deng2025cpretdatasetbenchmarkmodel}, Claude 3.7 Sonnet \cite{zheng2025livecodebenchproolympiadmedalists, deng2025cpretdatasetbenchmarkmodel}, Qwen models \cite{zheng2025livecodebenchproolympiadmedalists, jain2024livecodebenchholisticcontaminationfree, yu2024humanevalprombpppro, quan2025codeelobenchmarkingcompetitionlevelcode}, GPT-4-omni \cite{zheng2025livecodebenchproolympiadmedalists, jain2024livecodebenchholisticcontaminationfree, hossain2025llmprosanalyzinglargelanguage, yu2024humanevalprombpppro, quan2025codeelobenchmarkingcompetitionlevelcode, deng2025cpretdatasetbenchmarkmodel}, and GPT-3.5 Turbo \cite{shi2024languagemodelssolveolympiad, jain2024livecodebenchholisticcontaminationfree}. This section provides information about the models used:

\begin{table}[h]
\centering
\caption{Overview of models with organization, context window and cut-off date.}
\label{tab:models}
\begin{tabular}{p{4.4cm} p{1.5cm} r p{2.85cm}}
\toprule
\textbf{Full Model Name} & \textbf{Org.} & \textbf{Context Tokens} & \textbf{Cut-off Date} \\
\midrule
DeepSeek-R1 \cite{deepseek2025deepseekr1} & DeepSeek & 131K & Not specified \\
Gemini 2.0 Flash \cite{deepmind2025gemini2flash} & Google & 1 million & June 2024 \\
Claude 3.7 Sonnet \cite{anthropic2025claude37sonnet} & Anthropic & 200K & October 2024 \\
Qwen2.5-Coder-32B \cite{alibaba2025qwenmax} & Alibaba & 128K & End of 2023 \\
GPT-4-omni \cite{openai2024gpt4ocard} & OpenAI & 128K & September 2023 \\
GPT-3.5 Turbo \cite{openai2024gpt35turbo} & OpenAI & 16K & September 2021 \\
\bottomrule
\end{tabular}
\end{table}

Another decisive factor in the model selection was the diversity in training focus and knowledge cutoff date: we included a code-specialized model (Qwen2.5-Coder-32B), a model known for advanced reasoning capabilities (DeepSeek-R1), a less powerful model (GPT-3.5 Turbo), as well as general-purpose models with broader training data.

\subsection{Information Retrieval Algorithm}
\label{subsec:ir}

The IR algorithm retrieves similar solved problems from the database. First, each problem description in the database is transformed into a high-dimensional vector embedding using the `all-MiniLM-L6-v2'\footnote{\url{https://huggingface.co/sentence-transformers/all-MiniLM-L6-v2}} model. These embeddings capture the semantic essence of the problems \cite{embeddings}. The collection of embeddings is then indexed using FAISS (Facebook AI Similarity Search) \cite{FAISS_Johnson2017}, employing an `IndexFlatL2' structure that computes Euclidean distances. Because the vectors are already normalized, we can compute the similarity directly from the Euclidean distance (on a 0-to-1 scale).

A distinctive aspect of our IR system is its adaptive threshold mechanism, introduced by \cite{ReEffIRDinamicoIntroduzido}. Instead of a fixed cutoff, the threshold for deeming a retrieved problem ``similar enough'' is dynamically adjusted based on the highest similarity score. The initial threshold used for the similarity is 70\%, but if the highest similarity is strong (above 85\%), the threshold becomes more stringent; if weaker (less than 75\%), it relaxes, but never below the 70\% minimum. Similar adaptive strategies are used in RAG to reduce noise \cite{IRthreshold}, and we apply it with the same intent. As an additional measure to mitigate noise, the system is configured to return a maximum of the three most similar problems.

\subsection{Dataset Overview}
\label{subsec:dataset}

We utilized the CodeContests dataset \cite{li2022competition} for this study, designating its training set as the knowledge base for our RAG system and its test set for evaluation. The dataset, sourced primarily from CodeForces problems, is structured as a JSON file where each entry contains the problem's title, its full description, input/output examples, the official CodeForces rating (when applicable), relevant thematic tags, and verified code solutions. These problems do not require a specific programming language for implementation, allowing for flexibility in the solution approaches. This section describes how we leveraged this dataset to construct our RAG pipeline and analyzes the distribution of difficulties and categories of the problems used.

\begin{figure}[!h]
\centering
\includegraphics[width=0.877\textwidth]{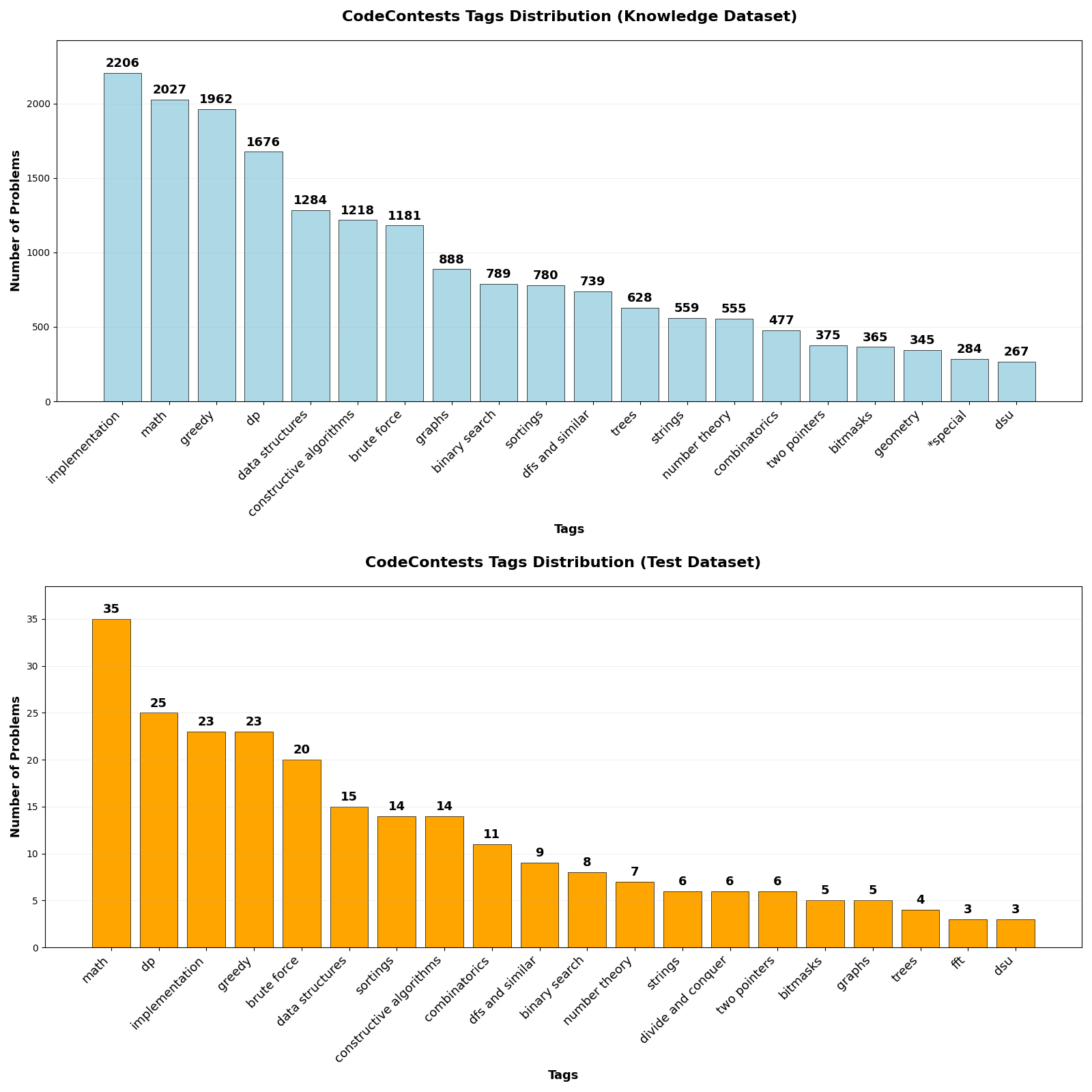}
\caption{Categories of the Problems from the Knowledge and Test Datasets}
\label{fig:categories}
\end{figure}

Of the 165 problems in the CodeContests test set, 81 had at least one similar problem found in the knowledge base by our Information Retrieval algorithm. The evaluation was therefore performed on this subset, as the remaining problems had no retrieved counterparts and would not receive any additional information from the RAG system. Furthermore, none of the problems in this evaluation set were part of the RAG’s knowledge base, ensuring a fair assessment of its generalization capabilities \cite{RAGIntroduction, gao2023retrieval, soares2025machine, miranda2023projeto}. Problems can be classified into one or more categories (e.g., dynamic programming, sorting, binary search), and the distribution of these thematic tags across both datasets is shown in Figure~\ref{fig:categories}.

To better understand the difficulty distribution of the problems, we performed a classification based on their CodeForces Elo ratings. While the Elo system provides a continuous scale, we grouped the problems into three discrete categories for a better analysis: Easy, Medium, and Hard. Problems with an Elo rating between 0 and 1000 were classified as Easy, those in the range (1000, 2000] as Medium, and any problem rated above 2000 as Hard. It is worth noting that while every problem in the test set is from CodeForces, the training set (our knowledge base) also contains problems from other sources, so not all of them had the CodeForces Elo information available. Table~\ref{tab:difficulty-distribution} shows the distribution of these difficulty levels across the knowledge base and our filtered test dataset. 

\begin{table}[!h]
\centering
\caption{Distribution of Problem Difficulty in the Knowledge and Test Datasets}
\label{tab:difficulty-distribution}
\begin{tabular}{l cc @{\hskip 0.2cm} cc}
\toprule
& \multicolumn{2}{c}{\textbf{Knowledge Dataset}} & \multicolumn{2}{c}{\textbf{Test Dataset}} \\
\cmidrule(r{0.4em}){2-3} \cmidrule(l{-0.1em}){4-5}
\textbf{Difficulty} & \textbf{Count} & \textbf{Proportion (\%)} & \textbf{Count} & \textbf{Proportion (\%)} \\
\midrule
Easy   & 901  & 11.70 & 18 & 22.22 \\
Medium & 3759  & 48.81 & 31 & 38.27 \\
Hard   & 3041  & 39.49 & 32 & 39.51 \\
\bottomrule
\end{tabular}
\end{table}

\subsection{Min‑K‑Based Membership Inference Analysis}

To complement the RAG evaluation and assess whether any code solutions in the test set were potentially present in the training corpora of the evaluated LLMs, we implemented a Min-K Membership Inference approach~\cite{r1}. This method estimates the likelihood that a given code sample was part of a model’s pretraining data, thereby quantifying exposure bias and memorization within large language models.

The core intuition behind this algorithm is that language models assign systematically higher confidence to samples encountered during training compared to unseen ones. Training data typically yield higher average token probabilities (or equivalently, lower perplexities). Rather than relying on mean likelihoods, the Min-K approach focuses on the lowest \( K\% \) of token log-probabilities, representing the model’s most uncertain predictions, because these regions provide stronger discriminatory signals between memorised and novel samples~\cite{r1}. By analyzing the extremal uncertainty pattern rather than overall fluency, the method enhances detection precision and reduces false positives.

Algorithm~\ref{alg:mink} summarizes the membership inference detection process for code snippets. Given a tokenized code sequence \( x = (x_1, x_2, \ldots, x_n) \), the model \( f_{\theta} \) computes the conditional log-probabilities for each token. These values are sorted in ascending order to extract the \textit{Min-K\%} subset, from which a stability-based membership score is computed as:

\[
S_{\text{MinK}}(x) = \frac{1}{K} \sum_{i=1}^{K} \text{min-logprob}(x_i)
\]

If this score exceeds a threshold \( \tau \), the sample is classified as Non-member (likely unseen by the model); otherwise, it is labeled as Member, suggesting partial or full memorization.

\begin{algorithm}[H]
\caption{Membership Inference Detection for Code Snippets}
\label{alg:mink}
\KwIn{Token sequence \( x = (x_1, x_2, \ldots, x_n) \); threshold \( \tau \); parameter \( K\% \)}
\KwOut{Membership classification (Member / Non-member)}

\For{\(i = 1 \text{ to } n\)}{
    Compute log-probability \( p(x_i | x_1, \ldots, x_{i-1}) \) via model \( f_\theta \)\;
}
Sort log-probabilities in ascending order \( L_{\text{sorted}}(x) \)\;
Extract the lowest \( K\% \) values: \( \text{MinK}(x) = L_{\text{sorted}}(x)[1:K] \)\;
Compute membership score: \( S_{\text{MinK}}(x) = \frac{1}{K} \sum_{i=1}^{K} \text{MinK}(x)[i] \)\;
\If{\( S_{\text{MinK}}(x) \geq \tau \)}{
    \Return \textbf{Non-member} (code likely unseen)\;
}
\Else{
    \Return \textbf{Member} (code likely in training data)\;
}
\end{algorithm}

To transform continuous Min-K scores into binary membership classifications, we apply a percentile-based thresholding strategy~\cite{r2}. The threshold \( \tau \) is defined as the average of the 75th percentile of known member scores and the 25th percentile of non-member scores. This empirically driven approach efficiently separates the two distributions while maintaining robustness to outliers and noise. Compared to fixed or heuristic thresholds, percentile-based selection adapts better to model variance and dataset composition, leading to more stable and interpretable decision boundaries.
The overall inference workflow comprises three main phases:

\begin{enumerate}

    \item \textit{Problem Evaluation Phase:} For each competitive programming problem, the corresponding verified solution code is tokenized and evaluated by the target LLM. In this phase, we considered the best solution for each problem (a total of 81 problems) across the six selected LLMs. Owing to the high computational and API costs associated with probability estimation, only one representative solution per problem was analyzed. The best solution was determined according to the shortest and earliest correct submission among all generated outputs. This choice ensures consistency across models while maintaining a fair and cost-efficient evaluation process.
    
    \item \textit{Detection Phase:} The Min-K algorithm computes per-token log-probabilities, sorts them, and derives \( S_{\text{MinK}}(x) \) for each solution. This score is then compared against \( \tau \) to classify each code sample as Member or Non-member.
    
    \item \textit{Analysis Phase:} Membership inference outcomes are aggregated to estimate model exposure levels across problem categories and difficulty levels, complementing the RAG-based performance evaluation.
    
\end{enumerate}

By integrating this step, our methodology not only measures performance improvement via retrieval augmentation but also accounts for potential memorization biases in LLMs. This enriches the interpretability of results, distinguishing genuine reasoning gains from possible recall of pretraining data in competitive programming contexts.

\subsection{Experiments}
\label{subsec:experiments}

To perform the experiment, we designed an experiment involving the six selected LLMs and the 81 problems from the filtered test set. Each problem was presented to every LLM under three distinct conditions. The first scenario, a zero-shot baseline, provided the LLM with only the original problem description. The second, a random-retrieval baseline, augmented the prompt with randomly selected problems from the knowledge base to analyze the effect of simply adding more context. The third scenario, our proposed RAG approach, enhanced the prompt with the most similar problems identified by our IR algorithm. For the RAG scenario, each test problem was matched against all problems in the knowledge base using the IR system.

The semantic similarity was computed exclusively using the textual content of the problems' descriptions. This was a choice to simulate a scenario where, upon encountering a new problem, one only has access to its textual statement, not pre-existing metadata like a solution or tags. After generation, all outputs were submitted to the CodeForces judge system, which verifies correctness by running the solution against a comprehensive set of hidden test cases and returns whether the solution was accepted or rejected (rejection occurs if any test case fails).

The structure of the prompts was kept consistent in all experiments to ensure a fair comparison. For the zero-shot baseline, the prompt contained the problem's title, its full textual description, and any provided input/output examples, followed by an instruction to generate a solution. For the few-shot conditions (random and RAG-based), the prompt began with the target problem's information, followed by a clear demarcation for the upcoming examples. Each retrieved example included its original title, description, tags, similarity score, and verified solution. In the prompts that had examples, an additional instruction was included to direct the model to take advantage of the provided examples when formulating its solution.

The primary objective was to quantify the performance improvement attributable to the quality of the examples. We measured performance using two metrics: the overall pass rate (pass@1), calculated as the percentage of problems correctly solved on the first attempt, and the Elo rating of each model --- which is translated from the Elo data of the solved and unsolved problems. The pass rate provides a direct measure of success, while analyzing the Elo ratings offers a nuanced understanding of the model's capability to solve problems.

In addition to these benchmarks, we performed a complementary Min‑K Membership Inference analysis on the generated code solutions. This step estimated the likelihood that a given solution resembled or originated from the model’s pretraining data. We calculated a probability confidence score for each line of code within a problem’s best solution, indicating its likelihood of being a member (i.e., part of the model’s pretraining data) or a non-member, as determined by Algorithm \ref{alg:mink}. 


\section{Results and Analysis}
\label{sec:results}

This section presents the results of our experiments, analyzing the performance of six LLMs on 81 problems across three prompting conditions: zero-shot, random-retrieval, and our RAG approach. We evaluated performance using two primary metrics: the pass@1 rate and the Elo rating. Other factors like the categories and difficulty of the problems are also considered. 

On average, our RAG approach achieved a 20.37\% success rate, outperforming both the zero-shot (17.9\%) and random-retrieval (17.28\%) approaches. These results demonstrate a crucial trade-off. On the one hand, the retrieval of any additional problem inevitably introduces a degree of noise to the prompt \cite{havrilla2024understandingeffectnoisellm}. On the other, selecting the right examples can provide useful content and insights to the model \cite{brown2020languagemodelsfewshotlearners}. The performance decrease seen with random retrieval (-0.62\%) suggests that when examples are irrelevant, the negative impact of this noise reduces the model's reasoning process. However, the 2.47 percentage point gain from our RAG method demonstrates that when the examples are contextually relevant, the value of the provided insights outweighs this negative impact, leading to performance gains.

\begin{figure}[!h]
\centering
\includegraphics[width=1    \textwidth]{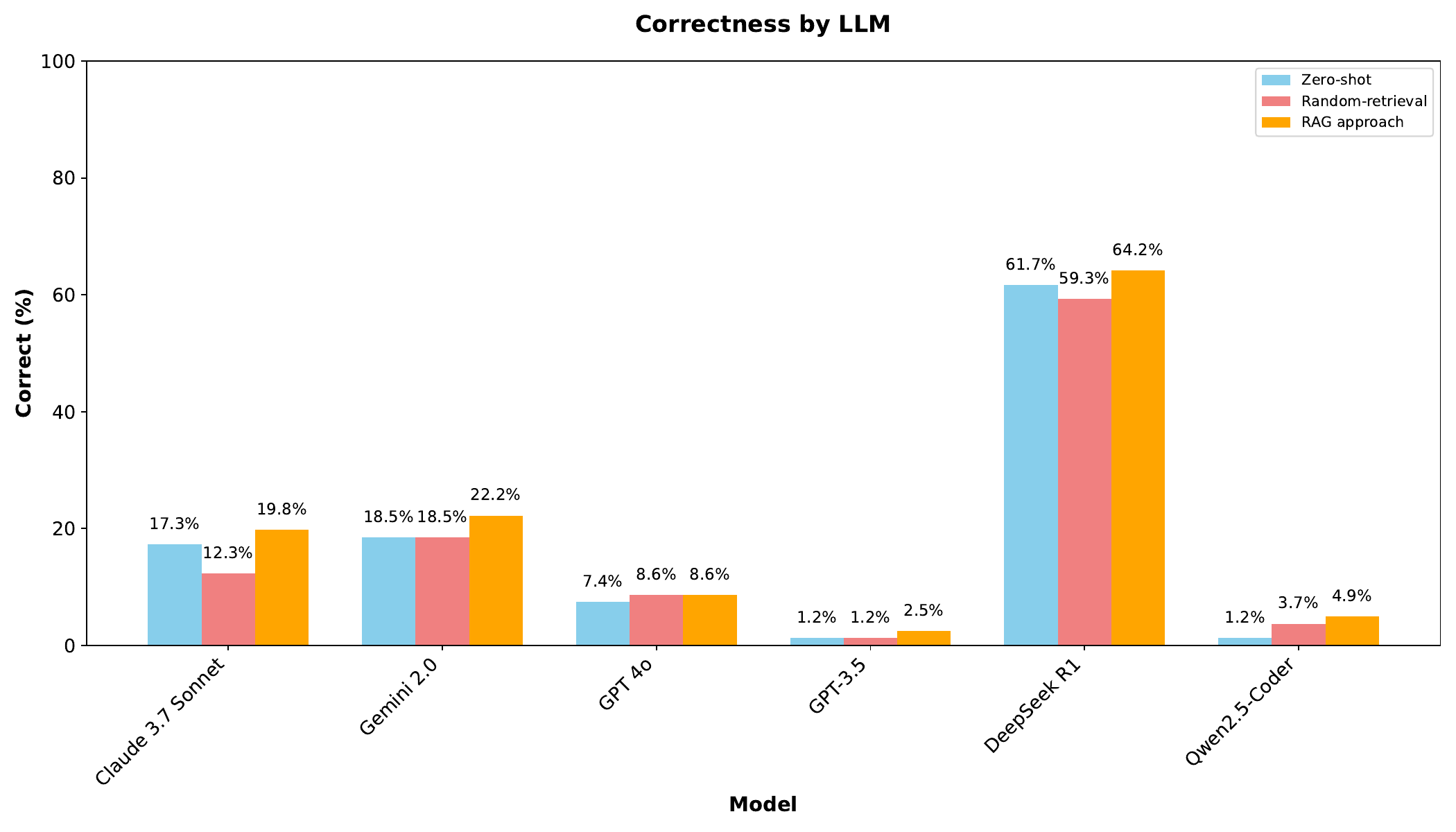}
\caption{Correctness analysis}
\label{fig:correctness}
\end{figure}

The individual pass@1 rates are detailed in Figure \ref{fig:correctness}. Weaker models like GPT-3.5 Turbo and Qwen2.5-Coder-32B struggled across all conditions. Although providing extra context led to some performance gains, the inherent limitations of these models were a more significant factor than the content of the examples. On the other side, DeepSeek-R1 had the best performance. Using our proposed framework, its pass rate improved from 61.7\% to 64.2\% (a relative improvement of 4.05\%). In this case, the random-retrieval scenario had the worst performance, with only 59.3\% correctness. Opposed to that, other models like GPT-4o and Qwen-2.5-Coder-32B showed an increase in accuracy in the random-retrieval scenario.

\begin{table}[ht]
\centering
\caption{Performance Based on Difficulty and Community Percentile Rating}
\label{tab:model_percentile}
\begin{tabular}{@{}lrrrrr@{}}
\toprule
\textbf{Model} & \textbf{Easy} & \textbf{Medium} & \textbf{Hard} & \textbf{Rating} & \textbf{Pct.} \\
\midrule
\textit{Zero-shot} & & & & & \\
DeepSeek R1       & 100.0\%& 80.6\% & 21.9\% & $2080$ & 2.9\% \\
Gemini 2.0        & 61.1\% & 12.9\% & 0.0\%  & $983$  & 54.3\% \\
Claude 3.7 Sonnet & 50.0\% & 16.1\% & 0.0\%  & $955$  & 57.0\% \\
GPT 4o            & 27.8\% & 3.2\%  & 0.0\%  & $706$  & 79.8\% \\
GPT-3.5           & 5.6\%  & 0.0\%  & 0.0\%  & $412$ & 90.3\% \\
Qwen2.5-Coder     & 0.0\%  & 3.2\%  & 0.0\%  & $412$ & 90.3\% \\
\midrule
\textit{Random-retrieval} & & & & & \\
DeepSeek R1       & 100.0\%& 77.4\% & 18.8\% & $2020$ & 3.3\% \\
Gemini 2.0        & 61.1\% & 12.9\% & 0.0\%  & $983$  & 54.3\% \\
Claude 3.7 Sonnet & 38.9\% & 9.7\%  & 0.0\%  & $839$  & 68.7\% \\
GPT 4o            & 33.3\% & 3.2\%  & 0.0\%  & $742$  & 77.5\% \\
Qwen2.5-Coder     & 11.1\% & 3.2\%  & 0.0\%  & $568$  & 87.3\% \\
GPT-3.5           & 5.6\%  & 0.0\%  & 0.0\%  & $412$ & 90.3\% \\
\midrule
\textit{RAG approach} & & & & & \\
DeepSeek R1       & 100.0\%& 83.9\% & 25.0\% & $2140$ & 2.2\% \\
Gemini 2.0        & 72.2\% & 16.1\% & 0.0\%  & $1069$ & 46.5\% \\
Claude 3.7 Sonnet & 50.0\% & 22.6\% & 0.0\%  & $1012$ & 51.7\% \\
GPT 4o            & 22.2\% & 9.7\%  & 0.0\%  & $742$  & 77.5\% \\
Qwen2.5-Coder     & 16.7\% & 3.2\%  & 0.0\%  & $621$  & 83.9\% \\
GPT-3.5           & 5.6\%  & 3.2\%  & 0.0\%  & $502$ & 88.0\% \\
\bottomrule
\end{tabular}
\end{table}

See Table \ref{tab:model_percentile} for Elo and difficulty analysis. These ratings were calculated using a MAP Elo estimation, and we collected its prior distribution's mean ($\mu_0$) and variance ($\sigma_0^2$) from CodeForces list of all users\footnote{https://codeforces.com/api/user.ratedList?activeOnly=false}. The percentile column represents the model's standing within the entire community. On average, our RAG method resulted in an Elo rating of 1014, an increase of 89 points over the zero-shot baseline (925) and 87 points over the random-retrieval baseline (927).

\begin{table}[ht]
\centering
\caption{DeepSeek-R1 Elo Rating and Rank by Problem Category}
\label{tab:category_performance}
\begin{tabular}{@{}lccc@{}}
\toprule
\textbf{Category} & \textbf{Zero-shot} & \textbf{Random-retrieval} & \textbf{RAG} \\
\midrule
math & 2177 (1.7\%) & 2042 (3.2\%) & 2177 (1.7\%) \\
dp & 2066 (3.0\%) & 1852 (5.4\%) & 2066 (3.0\%) \\
implementation & 1832 (5.7\%) & 1832 (5.7\%) & 1832 (5.7\%) \\
greedy & 1963 (3.9\%) & 1806 (6.1\%) & 2052 (3.1\%) \\
brute force & 1967 (3.8\%) & 1967 (3.8\%) & 1967 (3.8\%) \\
data structures & 2063 (3.0\%) & 1957 (3.9\%) & 1851 (5.5\%) \\
sortings & 1862 (5.3\%) & 1862 (5.3\%) & 1862 (5.3\%) \\
constructive algorithms & 1894 (5.0\%) & 1778 (6.6\%) & 1778 (6.6\%) \\
combinatorics & 2041 (3.2\%) & 2041 (3.2\%) & 1858 (5.4\%) \\
dfs and similar & 1915 (4.6\%) & 1915 (4.6\%) & 2075 (3.0\%) \\
binary search & 1442 (19.4\%) & 1609 (12.3\%) & 1609 (12.3\%) \\
number theory & 1604 (12.6\%) & 1604 (12.6\%) & 2191 (1.6\%) \\
strings & 1834 (5.7\%) & 1834 (5.7\%) & 2138 (2.2\%) \\
divide and conquer & 1859 (5.4\%) & 1859 (5.4\%) & 1859 (5.4\%) \\
two pointers & 2155 (2.0\%) & 2155 (2.0\%) & 2155 (2.0\%) \\
\bottomrule
\end{tabular}
\end{table}

To understand whether our RAG approach favors certain problem types, we analyzed the performance of the top-performing model, DeepSeek-R1, in different problem categories. Table \ref{tab:category_performance} shows the Elo rating (and the corresponding community rank) that the model achieved when evaluated on problems with a specific tag. 

The biggest difference was observed in problems tagged with `number theory', where the model's Elo increased from 1604 in both the zero-shot and random-retrieval scenarios to 2191 with RAG. This substantial leap is likely due to the logic-heavy nature of this category \cite{zheng2025livecodebenchproolympiadmedalists}. Number theory problems often hinge on discovering a mathematical insight that transforms a brute-force approach, such as an O(n) calculation, into an elegant O(1) formula. In these cases, a relevant example that reveals this insight can guide the model directly to the optimal solution. However, in categories such as `data structures' and `combinatorics', the performance decreased, suggesting that the retrieved examples can still introduce misleading information or noise.

\begin{figure}[!h]
\centering
\includegraphics[width=0.95    \textwidth]{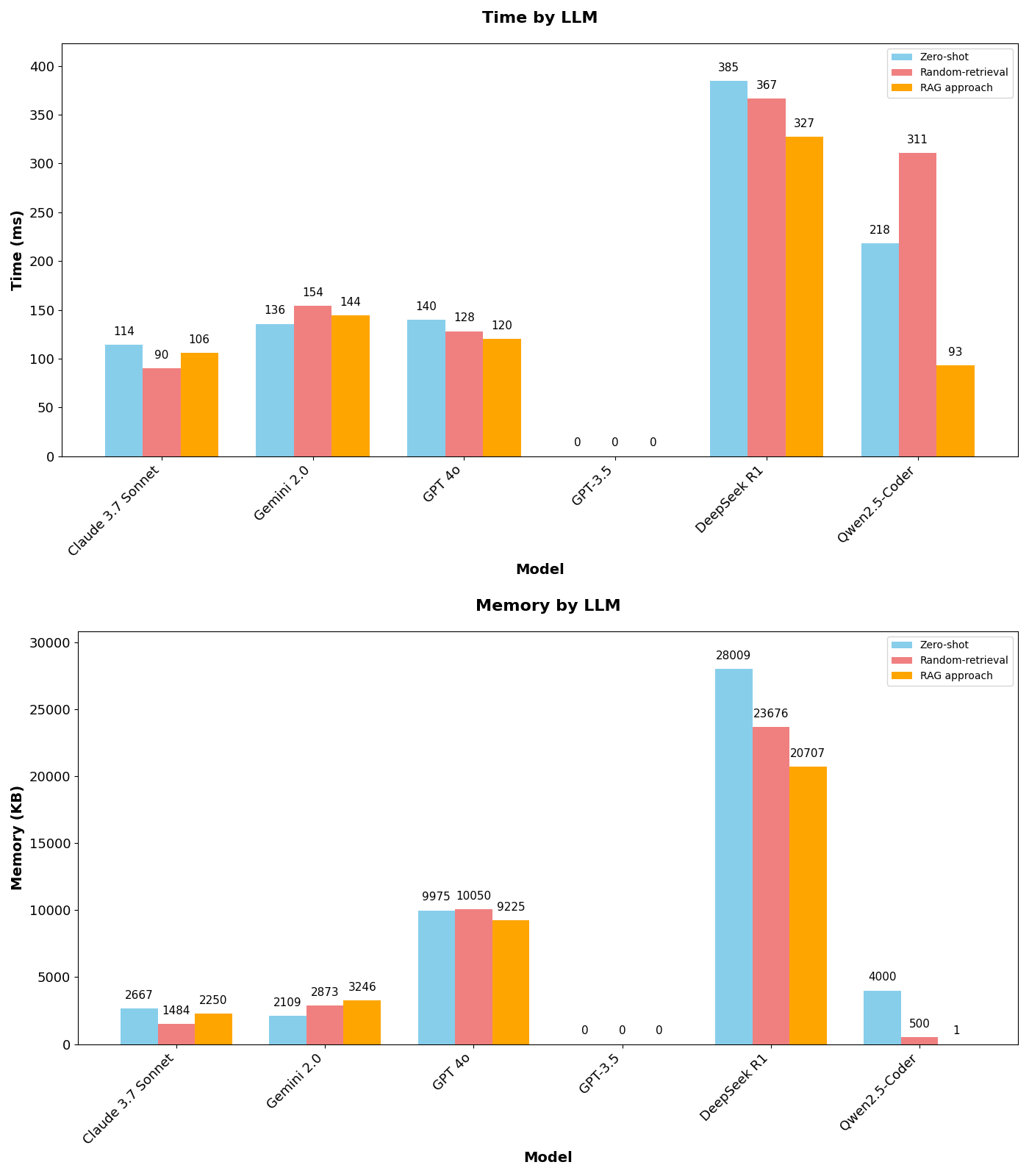}
\caption{Correctness analysis}
\label{fig:timememory}
\end{figure}

In the Figure \ref{fig:timememory}, there is a display of the average time and memory consumed in the runtime of the code generated for each problem by each LLM in all 3 scenarios. The problems considered were the ones that the given LLM solved in all the 3 approaches. Considering all models, in our RAG approach there was a decrease in computational resources used running the code with the test cases from the online judge.

For the 81 best solutions analyzed across the six evaluated LLMs DeepSeek‑R1, Gemini 2.0, Claude 3.7 Sonnet, GPT‑4o, Qwen2.5‑Coder‑32B, and GPT‑3.5 Turbo, the majority of code lines exhibited a membership confidence greater than 0.5. This indicates that a substantial portion of the generated code segments were likely seen or closely related to samples present in the models’ pretraining corpora. This finding suggests that earlier versions of these models were likely trained, at least partially, on subsets of the CodeContests dataset or similar competitive programming sources. However, it is plausible that these models were not trained on explicit \textless\textit{Problem}, \textit{Solution}\textgreater (i.e., input/output) pairs. Consequently, while they may have encountered fragments of relevant code during pretraining, they did not necessarily learn the mapping between specific problem statements and their corresponding solutions. This lack of structured alignment may explain why these models often struggle with complex or higher‑difficulty problems despite apparent familiarity with source code patterns.  

In contrast, more recent state‑of‑the‑art models have undergone advanced post‑training procedures such as reinforcement learning from human feedback (RLHF) and supervised fine‑tuning on curated \textless\textit{Problem}, \textit{Solution}\textgreater pairs across domains such as programming, mathematics, and logical reasoning \cite{r3}. These additional training stages strengthen their capacity for algorithmic reasoning and structured code synthesis, enabling them to generalize more effectively and solve harder problems. In future work, we plan to extend our analysis to the next generation of LLMs, such as GPT‑5, Gemini 3, and other emerging architectures. Evaluating these models under the same retrieval‑augmented and membership inference framework will allow a deeper understanding of how continual advances in model alignment and post‑training optimization contribute to reasoning ability and code generation performance in competitive programming tasks.

\section{Conclusion and Future Work}
\label{sec:conclusion}

This paper investigated whether Retrieval-Augmented Generation (RAG) can systematically improve Large Language Models (LLMs) on competitive programming — a setting where success depends on algorithmic reasoning, robustness to hidden test cases, and the ability to avoid hallucinated or logically inconsistent code. We proposed a similarity-based RAG pipeline that retrieves up to three relevant solved problems from a large knowledge base using embedding similarity and a dynamic, query-dependent threshold designed to reduce retrieval noise while preserving useful exemplars. We evaluated six representative LLMs on a filtered subset of 81 CodeContests test problems for which at least one similar training example could be retrieved, comparing three conditions: zero-shot, random-retrieval augmentation, and similarity-based RAG. 

Across models, similarity-based retrieval delivered consistent net gains over both baselines. On average, our RAG condition increased pass@1 to 20.37\%, compared to 17.9\% (zero-shot) and 17.28\% (random retrieval), indicating that example relevance is critical: adding unrelated context tends to introduce harmful noise, whereas aligned examples provide actionable algorithmic structure that outweighs this cost. In Elo terms, the proposed RAG approach produced an average rating of 1014, improving over the zero-shot (925) and random-retrieval (927) conditions, reinforcing that RAG benefits are not only about solving more problems but also about solving harder ones. 

Model-level analysis showed that stronger reasoners benefited most reliably from retrieval. DeepSeek-R1 achieved the highest overall performance and improved from 61.7\% to 64.2\% pass@1 under RAG, while random retrieval reduced it to 59.3\%, again highlighting the importance of retrieval precision. Category-level results suggested that RAG can be especially valuable in domains where a single key insight unlocks the solution (e.g., number theory, where DeepSeek-R1’s Elo increased markedly under RAG), while other categories exhibited occasional degradation, consistent with remaining sensitivity to misleading or weakly matched exemplars. 

To complement performance evaluation with an epistemic perspective, we applied a Min-K membership inference analysis to estimate whether generated solutions were likely influenced by memorized training patterns. The analysis indicated that a substantial portion of code lines across the best solutions had membership confidence above 0.5, suggesting meaningful exposure to competitive-programming-like code in model pretraining corpora. Importantly, this does not imply that models learned exact problem–solution mappings; rather, it supports the interpretation that models may have seen fragments of similar code patterns without robust alignment to the corresponding statements — helping explain why retrieval of structured, relevant worked examples can still provide measurable gains in problem-solving. 

Despite these findings, the study has limitations. First, evaluation was necessarily restricted to the subset of test problems with retrieved neighbors, which focuses the analysis on settings where retrieval is feasible. Second, we used a fixed embedding model and a bounded number of retrieved examples (up to three), leaving open questions about the optimal representation, retrieval depth, and prompt packaging under tight context budgets. Third, the membership inference is an indirect signal and should be interpreted as evidence of exposure patterns rather than definitive proof of duplication.

Future work will extend this framework along four directions. (i) Retrieval quality: explore hybrid retrieval (dense + sparse), reranking, and retrieval-time verification to further reduce harmful exemplars, especially for categories where RAG occasionally degrades performance. (ii) Adaptation and personalization: learn per-model and per-category retrieval policies (thresholds, number of examples, and formatting) to maximize gains while controlling noise. (iii) Broader benchmarks: replicate the pipeline on more recent and contamination-aware datasets, enabling a clearer separation between genuine reasoning improvements and latent memorization. (iv) Next-generation models and safety diagnostics: evaluate upcoming models under the same dual protocol—performance plus membership inference — to better understand how post-training advances (e.g., alignment and reasoning-tuned training) interact with retrieval grounding in competitive programming contexts.






\bibliographystyle{elsarticle-num} 
\bibliography{bibliography}

@inproceedings{r1,
title={Detecting Pretraining Data from Large Language Models},
author={Weijia Shi and Anirudh Ajith and Mengzhou Xia and Yangsibo Huang and Daogao Liu and Terra Blevins and Danqi Chen and Luke Zettlemoyer},
booktitle={The Twelfth International Conference on Learning Representations},
year={2024},
url={https://openreview.net/forum?id=zWqr3MQuNs}
}

@inproceedings{r2,
  title={Bayesian estimation of differential privacy},
  author={Zanella-Beguelin, Santiago and Wutschitz, Lukas and Tople, Shruti and Salem, Ahmed and R{\"u}hle, Victor and Paverd, Andrew and Naseri, Mohammad and K{\"o}pf, Boris and Jones, Daniel},
  booktitle={International Conference on Machine Learning},
  pages={40624--40636},
  year={2023},
  organization={PMLR}
}

@article{r3,
  title={Large Language Models Post-training: Surveying Techniques from Alignment to Reasoning},
  author={Tie, Guiyao and Zhao, Zeli and Song, Dingjie and Wei, Fuyang and Zhou, Rong and Dai, Yurou and Yin, Wen and Yang, Zhejian and Yan, Jiangyue and Su, Yao and others},
  journal={arXiv preprint arXiv:2503.06072},
  year={2025}
}

@INPROCEEDINGS{PerformanceLLMLeetCode,
  author={Wang, Lun and Shi, Chuanqi and Du, Shaoshuai and Tao, Yiyi and Shen, Yixian and Zheng, Hang and Qiu, Xinyu},
  booktitle={2024 4th International Symposium on Artificial Intelligence and Intelligent Manufacturing (AIIM)}, 
  title={Performance Review on LLM for solving leetcode problems}, 
  year={2024},
  pages={1050-1054},
}

@article{gao2023retrieval,
  title={Retrieval-augmented generation for large language models: A survey},
  author={Gao, Yunfan and Xiong, Yun and Gao, Xinyu and Jia, Kangxiang and Pan, Jinliu and Bi, Yuxi and Dai, Yixin and Sun, Jiawei and Wang, Haofen and Wang, Haofen},
  volume={2},
  pages={1},
  year={2023}
}

@inproceedings{LLMCodeGeneration,
  title={A review on code generation with llms: Application and evaluation},
  author={Wang, Jianxun and Chen, Yixiang},
  booktitle={2023 IEEE International Conference on Medical Artificial Intelligence (MedAI)},
  pages={284--289},
  year={2023},
  organization={IEEE}
}

@misc{hallucination,
      title={Exploring and Evaluating Hallucinations in LLM-Powered Code Generation}, 
      author={Fang Liu and Yang Liu and Lin Shi and Houkun Huang and Ruifeng Wang and Zhen Yang and Li Zhang and Zhongqi Li and Yuchi Ma},
      year={2024},
}

@misc{IRthreshold,
      title={RetrievalQA: Assessing Adaptive Retrieval-Augmented Generation for Short-form Open-Domain Question Answering}, 
      author={Zihan Zhang and Meng Fang and Ling Chen},
      year={2024},
}

@misc{BenchmarkLeetcode,
      title={LeetCodeDataset: A Temporal Dataset for Robust Evaluation and Efficient Training of Code LLMs}, 
      author={Xia, Yunhui and Shen, Wei and Wang, Yan and Liu, Jason Klein and Sun, Huifeng and Wu, Siyue and Hu, Jian and Xu, Xiaolong},
      year={2025},
}

@phdthesis{ReEffIRDinamicoIntroduzido,
author = {Balasubramanian, Niranjan},
advisor = {Allan, James},
title = {Query-dependent selection of retrieval alternatives},
year = {2011},
isbn = {9781267041692},
school = {University of Massachusetts Amherst},
}

@article{FAISS_Johnson2017,
  author    = {Jeff Johnson and Matthijs Douze and Herv{\'{e}} J{\'{e}}gou},
  title     = {Billion-Scale Similarity Search with GPUs},
  year      = {2017},
}

@misc{RAGIntroduction,
      title={Leveraging Passage Retrieval with Generative Models for Open Domain Question Answering}, 
      author={Gautier Izacard and Edouard Grave},
      year={2021},
      eprint={2007.01282},
}

@inproceedings{embeddings,
  title={Dependency-based word embeddings},
  author={Levy, Omer and Goldberg, Yoav},
  booktitle={Proceedings of the 52nd Annual Meeting of the Association for Computational Linguistics (Volume 2: Short Papers)},
  pages={302--308},
  year={2014}
}

@article{soares2025machine,
  title={Machine Learning-Assisted Tools for User eXperience Evaluation: A Systematic Mapping Study},
  author={Soares, Tales Schifelbein and Costa, Ricardo Luiz Hentges and Soares, Estefano and Calderon, Ivanilse and Lunardi, Gabriel Machado and Valle, Pedro Henrique Dias and Guedes, Gilleanes TA and Silva, Williamson},
  journal={Simp{\'o}sio Brasileiro de Sistemas de Informa{\c{c}}ao (SBSI)},
  pages={379--388},
  year={2025},
  publisher={SBC}
}

@inproceedings{miranda2023projeto,
  title={Projeto e avalia{\c{c}}{\~a}o de um template de worked examples para o ensino de programa{\c{c}}{\~a}o},
  author={Miranda, Andre LM and Garcia, Renato and Lunardi, Gabriel M and Vilela, Ricardo and Vale, Pedro HD and Silva, Williamson},
  booktitle={Simp{\'o}sio Brasileiro de Inform{\'a}tica na Educa{\c{c}}{\~a}o (SBIE)},
  pages={1673--1684},
  year={2023},
  organization={SBC}
}

@article{li2022competition,
  title={Competition-Level Code Generation with AlphaCode},
    author={Li, Yujia and Choi, David and Chung, Junyoung and Kushman, Nate and
    Schrittwieser, Julian and Leblond, R{\'e}mi and Eccles, Tom and
    Keeling, James and Gimeno, Felix and Dal Lago, Agustin and
    et al.},
  journal={arXiv preprint arXiv:2203.07814},
  year={2022}
}

@misc{zheng2025livecodebenchproolympiadmedalists,
      title={LiveCodeBench Pro: How Do Olympiad Medalists Judge LLMs in Competitive Programming?}, 
      author={Zihan Zheng and Zerui Cheng and Zeyu Shen and Shang Zhou and Kaiyuan Liu and Hansen He and Dongruixuan Li and Stanley Wei and Hangyi Hao and Jianzhu Yao and et al.},
      year={2025},
      eprint={2506.11928},
      archivePrefix={arXiv},
      primaryClass={cs.SE},
      url={https://arxiv.org/abs/2506.11928}, 
}

@misc{chen2021evaluatinglargelanguagemodels,
      title={Evaluating Large Language Models Trained on Code}, 
      author={Mark Chen and Jerry Tworek and Heewoo Jun and Qiming Yuan and Henrique Ponde de Oliveira Pinto and Jared Kaplan and Harri Edwards and Yuri Burda and Nicholas Joseph and et al.},
      year={2021},
      eprint={2107.03374},
      archivePrefix={arXiv},
      primaryClass={cs.LG},
      url={https://arxiv.org/abs/2107.03374}, 
}

@misc{yu2024humanevalprombpppro,
      title={HumanEval Pro and MBPP Pro: Evaluating Large Language Models on Self-invoking Code Generation}, 
      author={Zhaojian Yu and Yilun Zhao and Arman Cohan and Xiao-Ping Zhang},
      year={2024},
      eprint={2412.21199},
      archivePrefix={arXiv},
      primaryClass={cs.SE},
      url={https://arxiv.org/abs/2412.21199}, 
}

@misc{jimenez2024swebenchlanguagemodelsresolve,
      title={SWE-bench: Can Language Models Resolve Real-World GitHub Issues?}, 
      author={Carlos E. Jimenez and John Yang and Alexander Wettig and Shunyu Yao and Kexin Pei and Ofir Press and Karthik Narasimhan},
      year={2024},
      eprint={2310.06770},
      archivePrefix={arXiv},
      primaryClass={cs.CL},
      url={https://arxiv.org/abs/2310.06770}, 
}

@misc{hossain2025llmprosanalyzinglargelanguage,
      title={LLM-ProS: Analyzing Large Language Models' Performance in Competitive Problem Solving}, 
      author={Md Sifat Hossain and Anika Tabassum and Md. Fahim Arefin and Tarannum Shaila Zaman},
      year={2025},
      eprint={2502.04355},
      archivePrefix={arXiv},
      primaryClass={cs.CL},
      url={https://arxiv.org/abs/2502.04355}, 
}

@misc{deng2025cpretdatasetbenchmarkmodel,
      title={CPRet: A Dataset, Benchmark, and Model for Retrieval in Competitive Programming}, 
      author={Han Deng and Yuan Meng and Shixiang Tang and Wanli Ouyang and Xinzhu Ma},
      year={2025},
      eprint={2505.12925},
      archivePrefix={arXiv},
      primaryClass={cs.SE},
      url={https://arxiv.org/abs/2505.12925}, 
}

@misc{openai_prompt_engineering_2025,
  title        = {Prompt Engineering},
  howpublished = {\url{https://platform.openai.com/docs/guides/prompt-engineering}},
  year         = {2025},
  note         = {Acesso em: 26 de agosto de 2025}
}

@misc{deepseekai2025deepseekr1incentivizingreasoningcapability,
      title={DeepSeek-R1: Incentivizing Reasoning Capability in LLMs via Reinforcement Learning}, 
      author={DeepSeek-AI and Daya Guo and Dejian Yang and Haowei Zhang and Junxiao Song and Ruoyu Zhang and Runxin Xu and Qihao Zhu and Shirong Ma and Peiyi Wang and Xiao Bi and et al.},
      year={2025},
      eprint={2501.12948},
      archivePrefix={arXiv},
      primaryClass={cs.CL},
      url={https://arxiv.org/abs/2501.12948}, 
}

@misc{openai2024gpt4ocard,
      title={GPT-4o System Card}, 
      author={OpenAI and : and Aaron Hurst and Adam Lerer and Adam P. Goucher and Adam Perelman and Aditya Ramesh and Aidan Clark and AJ Ostrow and Akila Welihinda and Alan Hayes and Alec Radford and et al.},
      year={2024},
      eprint={2410.21276},
      archivePrefix={arXiv},
      primaryClass={cs.CL},
      url={https://arxiv.org/abs/2410.21276}, 
}

@techreport{anthropic2024claude35sonnet,
  title        = {{Model Card Addendum: Claude 3.5 Sonnet}},
  author       = {{Anthropic}},
  institution  = {Anthropic},
  year         = {2024},
  type         = {Model Card},
  url          = {https://www-cdn.anthropic.com/fed9cc193a14b84131812372d8d5857f8f304c52/Model_Card_Claude_3_Addendum.pdf}
}

@techreport{deepmind2025gemini25pro,
  title        = {{Model Card: Gemini 2.5 Pro}},
  author       = {{Google DeepMind}},
  institution  = {Google DeepMind},
  year         = {2025},
  type         = {Model Card},
  url          = {https://storage.googleapis.com/model-cards/documents/gemini-2.5-pro.pdf}
}

@misc{ma2023fairnessguidedfewshotpromptinglarge,
      title={Fairness-guided Few-shot Prompting for Large Language Models}, 
      author={Huan Ma and Changqing Zhang and Yatao Bian and Lemao Liu and Zhirui Zhang and Peilin Zhao and Shu Zhang and Huazhu Fu and Qinghua Hu and Bingzhe Wu},
      year={2023},
      eprint={2303.13217},
      archivePrefix={arXiv},
      primaryClass={cs.CL},
      url={https://arxiv.org/abs/2303.13217}, 
}

@misc{dickey2025failureplagiarismdetectioncompetitive,
      title={The Failure of Plagiarism Detection in Competitive Programming}, 
      author={Ethan Dickey},
      year={2025},
      eprint={2505.08244},
      archivePrefix={arXiv},
      primaryClass={cs.CY},
      url={https://arxiv.org/abs/2505.08244}, 
}

@misc{codeforces,
  author       = {Mike Mirzayanov},
  title        = {Codeforces},
  year         = {2010},
  url          = {https://codeforces.com/}
}

@misc{shi2024languagemodelssolveolympiad,
      title={Can Language Models Solve Olympiad Programming?}, 
      author={Quan Shi and Michael Tang and Karthik Narasimhan and Shunyu Yao},
      year={2024},
      eprint={2404.10952},
      archivePrefix={arXiv},
      primaryClass={cs.CL},
      url={https://arxiv.org/abs/2404.10952}, 
}

@misc{quan2025codeelobenchmarkingcompetitionlevelcode,
      title={CodeElo: Benchmarking Competition-level Code Generation of LLMs with Human-comparable Elo Ratings}, 
      author={Shanghaoran Quan and Jiaxi Yang and Bowen Yu and Bo Zheng and Dayiheng Liu and An Yang and Xuancheng Ren and Bofei Gao and Yibo Miao and Yunlong Feng and et al.},
      year={2025},
      eprint={2501.01257},
      archivePrefix={arXiv},
      primaryClass={cs.CL},
      url={https://arxiv.org/abs/2501.01257}, 
}

@misc{jain2024livecodebenchholisticcontaminationfree,
      title={LiveCodeBench: Holistic and Contamination Free Evaluation of Large Language Models for Code}, 
      author={Naman Jain and King Han and Alex Gu and Wen-Ding Li and Fanjia Yan and Tianjun Zhang and Sida Wang and Armando Solar-Lezama and Koushik Sen and Ion Stoica},
      year={2024},
      eprint={2403.07974},
      archivePrefix={arXiv},
      primaryClass={cs.SE},
      url={https://arxiv.org/abs/2403.07974}, 
}

@book{elo1978rating,
  author = {Elo, Arpad E. and Sloan, Sam},
  title = {The Rating of Chessplayers: Past and Present},
  year = {1978},
  note = {No Title}
}

@misc{zhou2024evaluatingsmoothcontrolattribute,
      title={Evaluating the Smooth Control of Attribute Intensity in Text Generation with LLMs}, 
      author={Shang Zhou and Feng Yao and Chengyu Dong and Zihan Wang and Jingbo Shang},
      year={2024},
      eprint={2406.04460},
      archivePrefix={arXiv},
      primaryClass={cs.CL},
      url={https://arxiv.org/abs/2406.04460}, 
}

@misc{openai2025introducing,
  author = {{OpenAI}},
  title = {Introducing OpenAI O3 and O4-mini},
  year = {2025},
  howpublished = {\url{https://openai.com/index/introducing-o3-and-o4-mini/}}
}

@misc{boubdir2023elouncoveredrobustnessbest,
      title={Elo Uncovered: Robustness and Best Practices in Language Model Evaluation}, 
      author={Meriem Boubdir and Edward Kim and Beyza Ermis and Sara Hooker and Marzieh Fadaee},
      year={2023},
      eprint={2311.17295},
      archivePrefix={arXiv},
      primaryClass={cs.CL},
      url={https://arxiv.org/abs/2311.17295}, 
}

@misc{deepseek2025deepseekr1,
  author = {DeepSeek AI},
  title = {DeepSeek-R1},
  year = {2025},
  howpublished = {\url{https://huggingface.co/deepseek-ai/DeepSeek-R1}}
}

@misc{deepmind2025gemini2flash,
  author = {{Google DeepMind}},
  title = {Model Card: {Gemini 2.0 Flash Reasoning}},
  year = {2025},
  howpublished = {\url{https://blog.google/technology/google-deepmind/gemini-model-updates-february-2025/}}
}

@misc{anthropic2025claude37sonnet,
  author = {Anthropic},
  title = {System Card: {Claude} 3.7 {Sonnet} (No Reasoning Variant)},
  year = {2025},
  howpublished = {\url{https://assets.anthropic.com/m/785e231869ea8b3b/original/claude-3-7-sonnet-system-card.pdf}}
}

@misc{alibaba2025qwenmax,
  author = {{Alibaba Cloud}},
  title = {Model Card: {Qwen} (Qwen 2.5-Coder-32B)},
  year = {2025},
  howpublished = {\url{https://huggingface.co/Qwen}}
}

@misc{openai2024gpt35turbo,
  author = {OpenAI},
  title = {{GPT-3.5 Turbo} Model Documentation},
  year = {2024},
  howpublished = {\url{https://platform.openai.com/docs/models/gpt-3.5-turbo}}
}

@misc{havrilla2024understandingeffectnoisellm,
      title={Understanding the Effect of Noise in LLM Training Data with Algorithmic Chains of Thought}, 
      author={Alex Havrilla and Maia Iyer},
      year={2024},
      eprint={2402.04004},
      archivePrefix={arXiv},
      primaryClass={cs.LG},
      url={https://arxiv.org/abs/2402.04004}, 
}

@misc{brown2020languagemodelsfewshotlearners,
      title={Language Models are Few-Shot Learners}, 
      author={Tom B. Brown and Benjamin Mann and Nick Ryder and Melanie Subbiah and Jared Kaplan and Prafulla Dhariwal and Arvind Neelakantan and Pranav Shyam and Girish Sastry and Amanda Askell and Sandhini Agarwal and Ariel Herbert-Voss and Gretchen Krueger and Tom Henighan and Rewon Child and Aditya Ramesh and Daniel M. Ziegler and Jeffrey Wu and Clemens Winter and Christopher Hesse and Mark Chen and Eric Sigler and Mateusz Litwin and Scott Gray and Benjamin Chess and Jack Clark and Christopher Berner and Sam McCandlish and Alec Radford and Ilya Sutskever and Dario Amodei},
      year={2020},
      eprint={2005.14165},
      archivePrefix={arXiv},
      primaryClass={cs.CL},
      url={https://arxiv.org/abs/2005.14165}, 
}



\end{document}